\documentclass[letterpaper]{article} 
\usepackage{aaai2027}  
\usepackage[hyphens]{url}  
\usepackage{graphicx} 
\usepackage{natbib}  
\usepackage{caption} 
\usepackage{algorithm}
\usepackage{algorithmic}
\usepackage{graphicx} 
\usepackage{natbib} 
\usepackage{caption} 
\usepackage{booktabs}
\usepackage{amsmath,amssymb}
\usepackage{array}
\usepackage{newfloat}
\usepackage{listings}
\DeclareCaptionStyle{ruled}{labelfont=normalfont,labelsep=colon,strut=off} 
\floatstyle{ruled}
\newfloat{listing}{tb}{lst}{}
\floatname{listing}{Listing}

\usepackage{booktabs}

\nocopyright
		
		\title{Selection-Invariant Communication Compilers for Privacy-Aware Multi-Agent LLM Workflows}
		\author{
			Jinghan Xu\textsuperscript{\rm 1},
			Longze Fan\textsuperscript{\rm 2},
			Zeyuan Wang\textsuperscript{\rm 3},
			Xinjin Li\textsuperscript{\rm 4},
			Hankai Liu\corresponding\textsuperscript{\rm 1}
		}
		\affiliations{
			\textsuperscript{\rm 1}Nankai University, China\\
			\textsuperscript{\rm 2}China University of Petroleum, China\\
			\textsuperscript{\rm 3}Sun Yat-sen University, China\\
			\textsuperscript{\rm 4}Columnbia University, USA\\
			
		}

\newcommand{\method}{SICC}
\newcommand{\gain}{ExcessGain}
\newcommand{\transcript}{T}
\newcommand{\outputy}{Y}
\newcommand{\priv}{S}
\newcommand{\req}{R}
\newcommand{\form}{F}
\newcommand{\opub}{O_{\mathrm{pub}}}
\newcommand{\apub}{A_{\mathrm{pub}}}
\newcommand{\xauth}{X_{\mathrm{auth}}}

\begin{document}
	\maketitle
	
	\begin{abstract}
		Structured multi-agent workflows exchange intermediate messages whose content and form can reveal private state even when the final output is safe. We identify \emph{selection-channel leakage}: after authorization fixes what may be released, a private-state-aware choice among semantically valid realizations creates an additional inference channel. We introduce the \textbf{\emph{selection-invariant communication compiler}}~(\method), which constrains this post-authorization representation kernel rather than prescribing templates. Any deterministic or independently public-randomized generator satisfying the invariant is valid; requirement-indexed canonical forms are one auditable implementation. We prove a compositional communication-layer guarantee: authorization, public-only form generation, and a dependency-safe utility gate make the emitted transcript reveal no information beyond the complete authorized view. Private-state-aware selection remains vulnerable after surface-disjoint and length-matched controls. Across 132 AgentLeak communication replays and 100 executable LangGraph tasks, deterministic \method{} retains complete protocol utility without a positive excess-gain signal; independent public randomization preserves the same result in AgentLeak and 480 controlled cases.

	\end{abstract}
	
	\section{Introduction}
	
	Multi-agent systems built on large language models (LLMs) increasingly coordinate through intermediate messages for task routing, delegation, and memory sharing. These transcripts create privacy risks that final-output auditing may miss: an external response can be safe while internal communication reveals private reasons or correlated operational facts. AgentLeak reports substantially higher leakage in inter-agent messages than in final outputs, while PAC-BENCH highlights the privacy--coordination trade-off in collaborative agents
	\citep{elyagoubi2026agentleak,park2026pacbench}.
	The resulting question is not only whether agents avoid explicit disclosure, but whether private state remains inferable from legitimate intermediate communication.
	
	Existing defenses address only part of this problem. Prompting and guarding are advisory, redaction may preserve correlated facts, and access control determines which information may cross a boundary rather than how an authorized release is expressed. Information-flow control (IFC) and declassification can model protected influence and intentional release
	\citep{denning1976lattice,sabelfeld2003language,costa2025fides}.
	However, once an authorized projection $\xauth$ is fixed, a sender may still choose among multiple valid realizations of the same release. If that choice depends on residual private state, the selected message form becomes an additional observable channel. We call this failure \emph{selection-channel leakage}.
	
	We address this post-authorization gap with the \emph{selection-invariant communication compiler} (\method{}). Let $\opub$ denote the attacker's pre-message observation and $\apub$ the independently specified schema of authorized fields. For a policy $\pi$ and $M\sim\pi(\cdot\mid X,\req,\outputy,\apub)$, we consider
	\begin{equation}
		\begin{aligned}
			\min_{\pi}\quad & I(\priv;M\mid\opub)\\
			\text{s.t.}\quad
			&\Pr[\mathrm{Succ}(M,\req)=1]\geq\tau,\\
			&\form(M)\perp\priv\mid\opub,\apub,\\
			&\operatorname{Fields}(M)
			\subseteq
			\operatorname{AllowedSlots}(\apub).
		\end{aligned}
	\end{equation}
	Equation~(1) states a policy-design objective rather than an optimization solved online by \method{}. The compiler constructs a certifiable feasible policy satisfying authorization, selection invariance, and protocol utility, without claiming global optimality over all utility-preserving generators.
	
	Selection invariance constrains the observable message-generation distribution rather than a particular renderer. A deterministic template, constrained decoder, or public-only randomized generator is admissible when its form does not depend on $\priv$ after conditioning on authenticated public inputs and policy, and when all instantiated values are authorized. Our implementation uses requirement-indexed canonical forms because their dependencies are directly auditable. It requires neither the evaluation label $\priv$ nor an attacker model at deployment. The \emph{Oracle Leakage-Ranked Selector} (OLRS) is used only as a nondeployable diagnostic: it ranks feasible messages with the evaluator label and demonstrates how per-message leakage minimization can itself induce form leakage.
	
	The guarantee is deliberately policy-relative. Access enforcement decides what may be released, whereas \method{} constrains how that authorized release is represented. It does not hide information already contained in the public or authorized view, repair an unsound authorization policy, or provide end-to-end secrecy. Its objective is to prevent additional inference caused by undeclared fields or private-state-dependent variation in message form.
	
	We evaluate \method{} on controlled enterprise, healthcare, and HR tasks, a 132-scenario AgentLeak communication-step replay, and an executable LangGraph workflow. Redaction and operational-fact sharing preserve protocol utility but yield $0.738\pm0.019$ excess attacker accuracy, while OLRS retains $0.225\pm0.019$ through selection and $0.207$ FormGain after semantic masking and exact length matching. Practical prompting and rewriting baselines also leak in the executable workflow, whereas fixed abstraction and receipt-only controls fail 39\% and 68\% of receiver requirements. Deterministic and public-randomized \method{} preserve complete
	protocol success on the evaluated AgentLeak and LangGraph cases
	with no statistically detectable positive excess gain under the
	evaluated attackers and schemas. A complementary human assessment
	also finds high receiver sufficiency with limited unnecessary
	disclosure.
	
	Our main contributions are:
	\begin{itemize}
		\item We identify and formalize \emph{selection-channel leakage}, in which private state is revealed through the sender's choice among individually sanitized or abstract message forms.
		
		\item We introduce \emph{selection invariance}, requiring the observable generation distribution to depend only on authenticated public inputs and authorization policy, rather than private state or private-state-dependent rankings.
		
		\item We instantiate this contract with auditable deterministic and public-randomized compilers and evaluate their privacy, utility, robustness, and workflow integration across controlled tasks, a public benchmark replay, and LangGraph execution.
	\end{itemize}

	\begin{figure*}[t]
		\centering
		\includegraphics[width=0.98\textwidth]{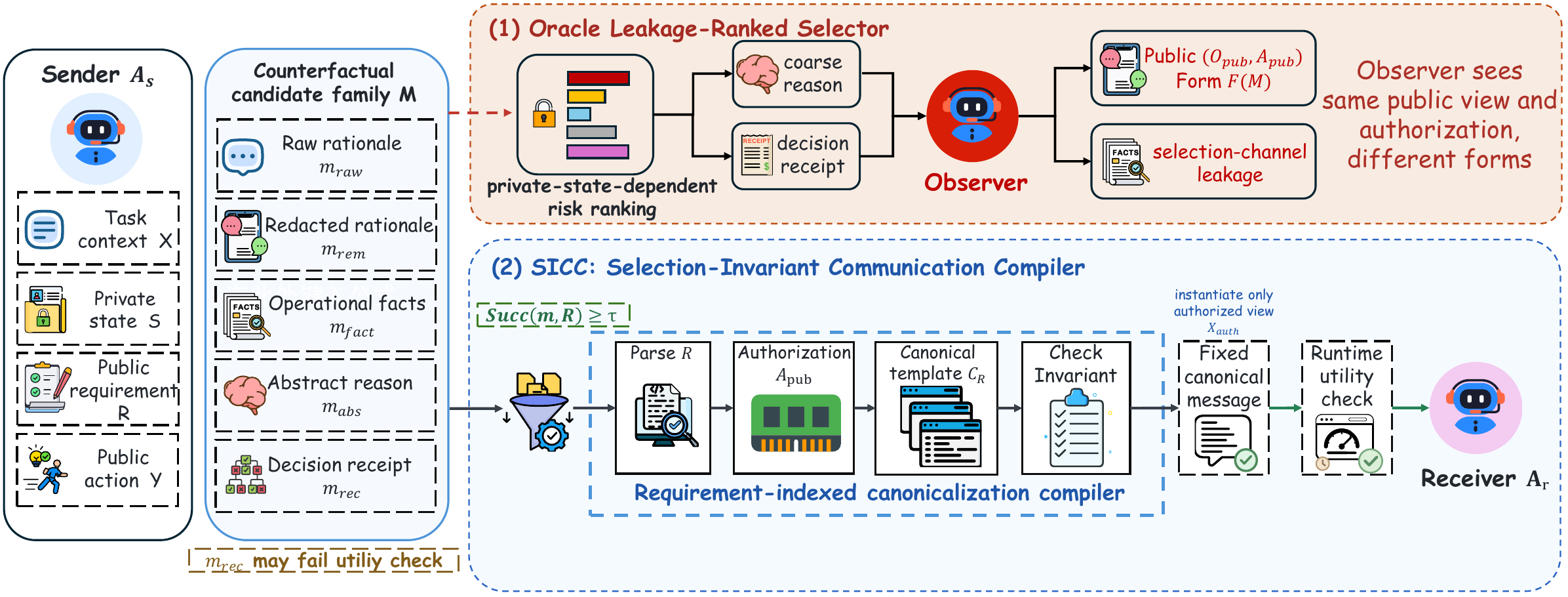}
		\caption{Selection-channel leakage versus SICC. \textbf{Top:} an oracle selector ranks valid messages using private-state-dependent scores, making the selected form observable. \textbf{Bottom:} SICC authenticates the receiver requirement, retrieves its authorization policy, projects authorized values from the sender context, fixes a requirement-indexed form, checks dependencies, and emits only after exact utility validation. Canonical templates are one auditable implementation of the compiler contract.}
		\label{fig:sicc-overview}
	\end{figure*}
	
	\section{Related Work}
	
	\paragraph{Contextual privacy and selective disclosure.}
	ConfAIde and PrivacyLens evaluate context-dependent disclosure norms, while AgentLeak, PAC-BENCH, POLAR-Bench, and CalBench study privacy leakage and privacy--coordination trade-offs in agent systems
	\citep{mireshghallah2024secret,shao2024privacylens,elyagoubi2026agentleak,park2026pacbench,zheng2026polarbench,zou2026calbench}.
	Related defenses provide step-specific privacy guidance or transform private context into task-sufficient abstractions
	\citep{wen2026contextualized,ngong2025protecting,li2025papillon,dou2024reducing}.
	These approaches primarily regulate communicated content; \method{} additionally constrains the observable policy that selects among valid message forms after authorization.
	
	\paragraph{Communication optimization and inference leakage.}
	Learned multi-agent protocols optimize communication through targeted channels or information-bottleneck objectives
	\citep{tishby2000information,das2019tarmac,wang2020efficient}.
	Such methods trade retained task information or communication cost against coordination performance, whereas \method{} constrains the distribution of the message-producing policy: even a short or individually sanitized message may leak when its form depends on $\priv$.
	Training-data extraction and PII inference instead recover memorized or identifying information from model outputs
	\citep{carlini2021extracting,kim2023propile}.
	Our attacker infers a case-level latent state from legitimate inter-agent transcripts after conditioning on the declared public view, separating communication-induced inference from model memorization.
	
	\paragraph{Information flow and choice-dependent side channels.}
	Noninterference, declassification, and quantitative information flow formalize protected influence, authorized release, and attacker knowledge gain
	\citep{denning1976lattice,sabelfeld2003language,sabelfeld2009declassification,smith2009foundations}.
	Recent agent systems such as Fides and CaMeL enforce complementary data-flow and capability boundaries
	\citep{costa2025fides,debenedetti2025camel}.
	General information-flow models can include message form in the observation space, and classical side-channel work establishes that observable choices may reveal protected state
	\citep{lampson1973confinement,kocher1996timing,kopf2007information}.
	Our narrower contribution is to isolate the post-authorization distribution over valid natural-language realizations, formalize selection invariance for inter-agent communication, and instantiate it as an executable compiler.
	
	\section{Problem Formulation}
	\label{sec:problem}
	
	\subsection{Task and Threat Model}
	
	We consider a sender agent $A_s$ and receiver agent $A_r$ collaborating on a task. The sender holds local context $X$, which may correlate with an evaluator-defined private state $\priv$, and the system produces output $\outputy$. The operational compiler receives $(X,\outputy,\req)$ and an authenticated policy catalog, but never $\priv$.
	
	Before communication, the attacker observes
	\begin{equation}
		\opub=(\outputy,\req,D,\ldots),
	\end{equation}
	where $D$ denotes any public domain or task-type information. The workflow retrieves an authorization object $\apub$, whose slot schema is fixed independently of the current case's $\priv$, and constructs the authorized value view
	\begin{equation}
		\xauth=
		\operatorname{PublicProject}_{\apub}(X;\req,\outputy).
	\end{equation}
	Here, $\opub$ is the attacker's pre-message view, $\apub$ specifies the permitted fields, and $\xauth$ contains their instantiated values. The authenticated requirement $\req$ may request, for example, a route, brief reason, or coordination fields; arbitrary receiver text cannot expand $\apub$ or the projection.
	
	The attacker knows the communication protocol and authorization schema, observes $\opub$ and optionally the transcript $\transcript$, and attempts to infer $\priv$. This captures a compromised receiver, orchestrator, logger, or later auditor without direct access to the private source. The label $\priv$ is used only for privacy evaluation and is unavailable to the deployed compiler.
	
	\paragraph{Trusted policy boundary.}
	The guarantee assumes a sound authorization boundary. Every value in $\xauth$ must belong to $\operatorname{AllowedSlots}(\apub)$, the projection must retain the fields required by the receiver, and the projection, allowlist, and generation policy must be constructed independently of the current $\priv$. Authorization determines what may be released; \method{} constrains its subsequent communication. Incorrectly authorized fields remain outside the guarantee.
	
	\subsection{Communication Leakage Channels}
	
	Transcript leakage has two sources. \emph{Content-channel leakage} occurs when a message contains explicit private labels or correlated facts. \emph{Selection-channel leakage} occurs when private context influences the observable message form, even if each candidate message is individually sanitized. For fixed $\opub$ and $\apub$, emitting a receipt in one case but a coarse rationale in another may therefore reveal $\priv$. We denote such observable form information by $\form(M)$, including message type, structure, or abstraction level.
	
	\subsection{Metrics}
	
	Our primary metric measures inference gain beyond the attacker's pre-message view:
	\begin{equation}
		\gain_{O}
		=
		\mathrm{Acc}(g(\opub,\transcript),\priv)
		-
		\mathrm{Acc}(g(\opub),\priv),
	\end{equation}
	where $g$ is an attacker evaluated on paired views with identical $\opub$. This accuracy difference is a computable proxy for $I(\priv;M\mid\opub)$, not an equality with mutual information. We retain the signed value for statistical analysis; a negative value indicates no evidence of positive excess leakage, rather than privacy improvement beyond $\opub$.
	
	We also measure leakage relative to the complete authorized view:
	\begin{equation}
		\begin{aligned}
			\gain_{A}
			={}&
			\mathrm{Acc}(g(\opub,\xauth,\transcript),\priv)\\
			&-
			\mathrm{Acc}(g(\opub,\xauth),\priv).
		\end{aligned}
	\end{equation}
	An authorized value not already contained in $\opub$ may increase $\gain_O$ while contributing no additional information beyond $\xauth$, yielding low $\gain_A$. In our controlled setting, the values emitted by \method{} are deterministic functions of $\opub$, so the two baselines coincide; their distinction nevertheless defines the scope of the policy-relative guarantee.
	
	Finally, we isolate form-only leakage with
	\begin{equation}
		\mathrm{FormGain}
		=
		\mathrm{Acc}(g(\opub,\form(M)),\priv)
		-
		\mathrm{Acc}(g(\opub),\priv).
	\end{equation}
	FormGain measures inference from observable form alone, whereas $\gain_O$ evaluates the complete transcript. Low FormGain does not imply low content leakage unless all instantiated fields satisfy the authorization policy.
	
	\section{Methodology}
	
	\subsection{Overview}
	
	\method{} is a compiler contract for message-producing policies rather than a particular template language. It combines authenticated policy retrieval, authorized projection, selection-invariant rendering, dependency checking, and exact protocol validation. Figure~\ref{fig:sicc-overview} contrasts a private-state-dependent selector, whose choice among valid messages may reveal private state, with \method{}, which determines message form from authenticated public inputs and instantiates only authorized fields. Our deterministic implementation uses requirement-indexed canonical forms for direct dependency auditing.

	\begin{table*}[t]
		\centering
		\small
		\caption{Running example for an HR sender whose private state is visa dependency. SICC removes undeclared content and private-state-dependent form selection, but does not hide authorized values.}
		\label{tab:running}
		\begin{tabular}{p{0.17\linewidth}p{0.47\linewidth}p{0.27\linewidth}}
			\toprule
			Policy & Message sent to receiver & Leakage mechanism \\
			\midrule
			Raw & Route requires HR priority because the employee has visa dependency and a renewal deadline. & Explicit private state. \\
			Redacted & Route requires HR priority because of renewal deadline, sponsor contact, and manager escalation. & Correlated facts reveal the state. \\
			Operational facts & Renewal deadline; sponsor contact; escalation queue. & Semantic-join leakage. \\
			OLRS & Sometimes a receipt, sometimes a coarse reason, depending on a private-state-aware risk score. & Selected form becomes a side channel. \\
			\method{} & Route requires HR coordination. No additional rationale is included. & Public-requirement-dependent form. \\
			\bottomrule
		\end{tabular}
	\end{table*}

	\subsection{Operational Compiler}
	
	Deployment follows
	\[
	(\req,\mathcal A)
	\rightarrow
	\apub
	\rightarrow
	\xauth
	\rightarrow
	G_{\req,\apub}
	\rightarrow
	\text{utility check}.
	\]
	The evaluated deterministic renderer performs neither candidate ranking nor attacker scoring. Its binary checker uses $\tau=1$; unmet requirements trigger abstention or escalation.
	
	\subsection{Selection Invariance and Guarantees}
	
	Selection invariance constrains the distribution of observable form rather than a specific rendering architecture. Let $U$ be public randomness sampled before private-state access, with
	\[
	U\perp(\priv,\xauth)\mid\opub,\apub,
	\]
	and let
	\[
	M\sim G_{\req,\apub}
	(\cdot\mid\opub,\xauth,U).
	\]
	Authorized values may fill declared slots, but the form-producing path must satisfy Definition~1. We evaluate deterministic and three-way public-randomized renderers; other generators are admissible only when their dependencies, fields, and utility can be certified.
	
	\begin{table}[t]
		\centering
		\scriptsize
		\caption{The deterministic renderer is indexed by authenticated requirements and policies, not private labels.}
		\label{tab:templates}
		\begin{tabular}{lll}
			\toprule
			Receiver requirement & Authorized fields & Canonical form \\
			\midrule
			Route only & domain, route & route receipt \\
			Brief reason & domain, route, coarse reason & fixed coarse rationale \\
			Coordination fields & domain, route, public slots & schema receipt \\
			\bottomrule
		\end{tabular}
	\end{table}
	
	\noindent\textbf{Definition 1 (selection-invariant policy).}
	Let $M\sim\pi(\cdot\mid X,\req,\outputy,\apub)$ and let $\form(M)$ denote its observable form. The policy is selection-invariant when
	\begin{equation}
		P(\form(M)\mid\priv,\opub,\apub)
		=
		P(\form(M)\mid\opub,\apub),
	\end{equation}
	equivalently,
	\[
	\form(M)\perp\priv\mid\opub,\apub.
	\]
	Any generator satisfying this invariant together with the authorized-field and utility constraints is a valid \method{} renderer. Deterministic templates with
	$\form(M)=h(\req,\apub)$ are one auditable realization.
	
	\noindent\textbf{Proposition 1 (selection-channel leakage).}
	Let $J$ denote the selected candidate index. If, for fixed $(\opub,\apub)$,
	\[
	\form(M)=\phi(J)
	\]
	and $\phi$ is injective on the selector's support, then
	\begin{equation}
		I(\priv;\form(M)\mid\opub,\apub)
		=
		I(\priv;J\mid\opub,\apub).
	\end{equation}
Hence, private-state-dependent candidate selection creates positive form-channel information whenever
\[
P(J\mid\priv,\opub,\apub)
\neq
P(J\mid\opub,\apub)
\]
on a set of positive probability.

\emph{Proof sketch.}
Under injectivity, $J$ and $\form(M)$ are recoverable from each other for each fixed public condition and therefore contain the same conditional information about $\priv$.

\noindent\textbf{Theorem 1 (compositional communication-layer noninterference).}
Represent an accepted message as $M=(\form,V)$, let $B$ be the utility checker's accept bit, and let the total observation $Z$ equal $M$ when $B=1$ and a fixed abstention token otherwise. Suppose:
\begin{enumerate}
	\item $V=v(\opub,\xauth,\req,\apub)$ and every field in $V$ belongs to $\operatorname{AllowedSlots}(\apub)$;
	\item $\form=f(\opub,\req,\apub,U)$ for
	$U\perp(\priv,\xauth)\mid\opub,\apub$ sampled before private-state access; and
	\item $B=q(\opub,\xauth,\req,\apub,U)$, with acceptance only when
	$\operatorname{Succ}(M,\req)=1$.
\end{enumerate}
Then
\begin{equation}
	I(\priv;Z\mid\opub,\xauth,\apub)=0.
\end{equation}
In particular, on accepted executions,
\begin{equation}
	I(\priv;M\mid\opub,\xauth,\apub)=0.
\end{equation}

\emph{Proof sketch.}
Conditioned on
$C=(\opub,\xauth,\apub)$,
the authorized values $V$ are fixed, while $(\form,B)$ depends only on $C$ and independent public randomness $U$. Thus
\[
I(\priv;\form,V,B\mid C)=0.
\]
Since $Z$ is a deterministic serialization of $(\form,V,B)$, the claim follows by data processing. The fixed abstention token prevents the accept/reject outcome itself from encoding residual private state.

Proposition~1 characterizes the positive selection channel, while Theorem~1 shows how authorization, form invariance, and a dependency-safe utility gate compose. Conditioning on $\xauth$ is essential: authorized values may themselves reveal $\priv$, so the result guarantees no additional information beyond the authorized view rather than absolute secrecy.

\noindent\textbf{Dependency enforcement.}
\textsc{CheckDependencies}$(G_{\req,\apub})$ restricts the form-producing path to authenticated public inputs and public randomness. Instantiated values may depend only on $(\opub,\xauth)$, and every emitted slot must belong to $\operatorname{AllowedSlots}(\apub)$. Violations reject compilation and trigger abstention or escalation.

The invariant concerns the renderer's dependencies, not fixed wording. Deterministic rendering makes those dependencies directly auditable, while public-randomized rendering shows that surface variation remains permissible when its selection is independent of private state. OLRS provides the contrasting diagnostic: individually sanitized candidates still leak when a private-state-dependent objective chooses their observable form.

\subsection{Privacy-Label-Free Deployment}

\method{} requires no per-message private-state labels, attacker models, provenance tags, or taint metadata. It composes with systems that authenticate receiver requirements and authorize data release: those systems establish the trusted boundary, while \method{} enforces the post-authorization communication contract.

\section{Experiments}
\label{sec:experiments}

\subsection{Setup}

\paragraph{Controlled scenarios.}
We construct programmatic tasks in enterprise procurement, healthcare referral, and HR support, with four evaluator-defined private states per domain. Each state has five hand-authored correlated facts, three of which are instantiated through fixed templates. The 12 states are balanced across three receiver requirements, and public routes are shared within each domain so that neither the route nor requirement directly identifies the private state. Training and test sets contain 2,400 and 900 cases with different seeds and identifiers but shared fact vocabulary. A separate surface-disjoint audit replaces all sentence scaffolds while retaining facts and labels; it tests robustness to phrasing rather than unseen semantics.

\paragraph{Baselines.}
We organize comparisons by purpose. \emph{OLRS} is an intentionally nondeployable diagnostic that ranks feasible raw, redacted, factual, abstract, and receipt messages using the evaluator label. Published practical defenses include Contextualized Privacy Defense (CPD) and Contextual Privacy Reformulation, evaluated through their released intervention interfaces
\citep{wen2026contextualized,ngong2025protecting}.
Generic controls include minimum-necessary prompting, semantic redaction, a privacy instructor, fixed abstraction without utility filtering, and receipt only. Proposed methods are deterministic and public-randomized \method{}, neither of which reads the private label or scores candidate leakage. An exact checker supplies the utility gate. All practical baselines receive the same sender context,
authenticated receiver requirement, and receiver-facing output
interface. They produce a single message, receive no access to the
evaluator-defined private label, and are evaluated without
test-label-dependent tuning. Implementation versions, prompts, and
workflow adaptations are reported in the appendix.

\paragraph{Evaluation protocol.}
We use held-out Qwen2.5-3B/7B and Llama-3-8B attackers on 900 cases and Mistral-7B on a 200-case subset. Receiver evaluation uses Qwen2.5-3B and Mistral-7B on 480 shared cases, requiring recovery of the route and any requested public reason or domain field. Metrics and paired differences use 10,000 case-level bootstrap samples; paired McNemar tests use Holm correction within each table family, with adjusted $p<0.001$ marked. All models run locally on RTX 3090 GPUs.

\subsection{Diagnosing Message-Form Leakage}

\begin{table}[t]
	\centering
	\footnotesize
	\caption{FormGain over 900 cases. Cells are paired-bootstrap 95\% CIs; $\dagger$ marks Holm-adjusted McNemar $p<0.001$ against \method{}. Masked messages contain 64 lexical tokens.}
	\label{tab:formonly}
	\setlength{\tabcolsep}{3pt}
	\begin{tabular}{lcc}
		\toprule
		Observable view & OLRS & \method{} \\
		\midrule
		Explicit form ID & $.207^{\dagger}$ [.180,.233] & .000 [.000,.000] \\
		Length only & $.243^{\dagger}$ [.216,.272] & .000 [.000,.000] \\
		Structure only & $.236^{\dagger}$ [.208,.263] & .000 [.000,.000] \\
		Masked + 64 tokens & $.207^{\dagger}$ [.180,.233] & .000 [.000,.000] \\
		\bottomrule
	\end{tabular}
\end{table}

Table~\ref{tab:formonly} isolates leakage from observable form. OLRS remains predictive when the attacker sees only length or nonlexical structural counts and slot indicators. FormGain also persists after masking all case-specific labels, facts, domains, routes, and coarse reasons and normalizing message length. In contrast, \method{} remains at the pre-message baseline because its form is not selected by private-state-aware ranking. These controls isolate the proposed channel, although unrestricted paraphrase families remain outside this audit.

\begin{table}[t]
	\centering
	\footnotesize
	\caption{Component ablation on 132 AgentLeak replays. A/I/G denote authorization, form invariance, and utility gating; P denotes partial authorization.}
	\label{tab:ablation}
	\setlength{\tabcolsep}{2pt}
	\begin{tabular}{lcccccc}
		\toprule
		Policy & A & I & G & Recv. & Extra & FormGain \\
		\midrule
		Fixed full msg. & P & Y & N & 1.00 & 1.00 & .000 [.000,.000] \\
		Authorized slots & Y & N & Y & 1.00 & .00 & .515 [.432,.598] \\
		Canonical form & Y & Y & N & .00 & .00 & .000 [.000,.000] \\
		Full \method{} & Y & Y & Y & 1.00 & .00 & .000 [.000,.000] \\
		\bottomrule
	\end{tabular}
\end{table}

Table~\ref{tab:ablation} removes one component at a time. Fixed full-context communication is invariant but exposes undeclared values; authorized fields alone remain vulnerable when a withheld-value-dependent selector chooses among equivalent forms; and canonical output without utility gating omits required fields. Only full \method{} jointly satisfies authorization, form invariance, and receiver completeness. The withheld-value hash is a controlled dependence intervention, not a semantic claim about AgentLeak labels.

\subsection{Transcript Leakage and Adaptive Attacks}

\begin{table*}[t]
	\centering
	\footnotesize
	\caption{One-shot leakage relative to $\opub$. Panel (a) reports the held-out lexical attacker; panel (b) reports ExcessGain for four held-out LLM attackers. Cells are point estimates with paired-bootstrap 95\% CIs.}
	\label{tab:main}
	\setlength{\tabcolsep}{0.5pt}
	\begin{tabular}{lccc@{\hspace{10pt}}lccc}
		\toprule
		\multicolumn{4}{c}{(a) Held-out lexical attacker} &
		\multicolumn{4}{c}{(b) LLM-attacker ExcessGain} \\
		Policy & Pre-message & Transcript & Excess &
		Attacker & Redacted & Operational & \method{} \\
		\midrule
		Redacted & .261 [.242,.280] & 1.000 [1.000,1.000] & .738 [.719,.757] &
		Qwen2.5-3B & .181 [.147,.216] & .159 [.130,.188] & .000 [.000,.000] \\
		Operational & .261 [.242,.280] & 1.000 [1.000,1.000] & .738 [.719,.757] &
		Qwen2.5-7B & .258 [.220,.293] & .280 [.248,.313] & $-.017$ [$-.040$,.007] \\
		Abstract & .261 [.242,.280] & .216 [.199,.233] & $-.046$ [$-.068$,$-.024$] &
		Llama-3-8B & .107 [.068,.146] & .194 [.154,.233] & .000 [$-.039$,.039] \\
		OLRS & .261 [.242,.280] & .487 [.460,.514] & .225 [.206,.244] &
		Mistral-7B & .260 [.195,.330] & .250 [.180,.320] & $-.045$ [$-.100$,.010] \\
		\method{} & .261 [.242,.280] & .216 [.199,.233] & $-.046$ [$-.068$,$-.024$] &
		& & & \\
		\bottomrule
	\end{tabular}
\end{table*}

Table~\ref{tab:main} shows that removing explicit labels does not eliminate inference from correlated facts. Redacted and operational messages have positive ExcessGain across held-out attackers. Minimum-necessary prompting obtains .120 [.045,.195] on the Mistral subset, while the privacy instructor obtains .015 [$-.045$,.080]. In the lexical audit, CPD and Contextual Privacy Reformulation preserve receiver completion but retain ExcessGain .167 and .356, respectively, providing practical content-level comparisons rather than selection-invariance guarantees.

Surface-disjoint scaffolds reduce redacted/operational ExcessGain from .738 [.719,.757] to .663 [.632,.694], while OLRS remains .222 [.194,.250] and \method{} remains .000 [.000,.000]. Because fact vocabulary is shared, this establishes surface robustness only. A separate 480-case semantic-transfer audit replaces each fact with a manually paired paraphrase from a disjoint lexicon. Raw facts yield ExcessGain .533 [.490,.577], paraphrases retain .308 [.258,.358], and \method{} yields .029 [$-.004$,.063]. This tests a fixed semantic map rather than open-domain transfer.

\begin{table}[t]
	\centering
	\small
	\caption{Four-round adaptive attack over 600 cases. Lower is better.}
	\label{tab:adaptive}
	\setlength{\tabcolsep}{2pt}
	\begin{tabular}{lcc}
		\toprule
		Policy & Attack@4 & Cum. ExcessGain \\
		\midrule
		Redacted & .770 & .490 \\
		Minimum necessary & .700 & .340 \\
		Semantic redaction & .690 & .330 \\
		Privacy instructor & .570 & .170 \\
		CPD & .560 & .160 \\
		Contextual reform. & .680 & .310 \\
		Fixed abstraction & .170 & .040 \\
		Receipt only & .080 & .010 \\
		\method{} & \textbf{.040} & \textbf{.000} \\
		\method{}-random & \textbf{.040} & \textbf{.000} \\
		\bottomrule
	\end{tabular}
\end{table}

We further evaluate a four-round adaptive attacker that generates each follow-up from the accumulated transcript and attempts private-state recovery in the final round. As shown in Table~\ref{tab:adaptive}, published and generic content defenses retain positive cumulative gain. Deterministic and randomized \method{} retain zero cumulative ExcessGain; their .040 recovery rate equals information already available in the public view rather than transcript-induced gain.

\subsection{Receiver Utility and Human Evaluation}

\begin{table*}[t]
	\centering
	\small
	\caption{Receiver completion and transcript leakage over 480 shared cases. Rule denotes exact protocol completion; higher is better except for ExcessGain.}
	\label{tab:e2e}
	\setlength{\tabcolsep}{7pt}
	\begin{tabular}{lcccc}
		\toprule
		Policy & Rule & Qwen & Mistral & ExcessGain \\
		\midrule
		Redacted & 1.000 & .692 & .921 & .456 \\
		Minimum necessary & 1.000 & .820 & .960 & .300 \\
		Semantic redaction & 1.000 & .800 & .950 & .300 \\
		Privacy instructor & 1.000 & .880 & .970 & .160 \\
		CPD & 1.000 & .900 & .980 & .167 \\
		Contextual reform. & 1.000 & .860 & .970 & .356 \\
		Fixed abstraction & .610 & .860 & .940 & .050 \\
		Receipt only & .338 & .904 & .906 & $-.044$ \\
		\method{} & 1.000 & \textbf{.946} & \textbf{1.000} & \textbf{$-.008$} \\
		\bottomrule
	\end{tabular}
\end{table*}

Table~\ref{tab:e2e} separates exact protocol completion from
model-based receiver execution. CPD achieves the strongest receiver
completion among the practical content-level defenses, but retains
positive ExcessGain (.167), while Contextual Privacy Reformulation
retains .356. \method{} obtains .946 completion under Qwen and 1.000
under Mistral with $-.008$ ExcessGain. Receipt-only communication
performs well under forced-choice receivers but satisfies only .338
of exact receiver requirements, showing that model-based completion
cannot replace the protocol checker.

\begin{table*}[h]
	\centering
	\small
	\caption{LangGraph traces over 100 held-out tasks. Requirement pass is higher-is-better; leakage and attack success are lower-is-better.}
	\label{tab:langgraph}
	\setlength{\tabcolsep}{8pt}
	\begin{tabular}{lccc}
		\toprule
		Policy & Requirement pass & ExcessGain & Attack Success@4 \\
		\midrule
		Redacted & 1.000 & .470 & .760 \\
		Minimum necessary & 1.000 & .300 & .690 \\
		Semantic redaction & 1.000 & .300 & .680 \\
		Privacy instructor & 1.000 & .160 & .560 \\
		CPD & 1.000 & .160 & .550 \\
		Contextual reform. & 1.000 & .310 & .670 \\
		Fixed abstraction & .610 & .040 & .160 \\
		Receipt only & .320 & .010 & .070 \\
		\method{} & \textbf{1.000} & \textbf{.000} & \textbf{.040} \\
		\method{}-random & \textbf{1.000} & \textbf{.000} & \textbf{.040} \\
		\bottomrule
	\end{tabular}
\end{table*}

\paragraph{Human receiver assessment.}
To complement rule-based and model-based metrics, two annotators
independently assessed all method outputs on the same 100 receiver
tasks for task sufficiency and unnecessary disclosure. \method{}
achieves .960 sufficiency and .040 unnecessary disclosure, compared
with .930/.200 for CPD and .380/.030 for receipt-only communication.
These results provide complementary evidence that \method{} preserves
task-relevant information without relying on indiscriminate
disclosure, whereas receipt-only communication reduces disclosure
primarily by omitting required information. Full annotation
instructions, agreement statistics, and per-method results appear in
Appendix.

\subsection{Public-Benchmark and Workflow Evaluation}

\begin{table}[t]
	\centering
	\footnotesize
	\caption{AgentLeak communication-step replay on 132 public scenarios. Recv.: all authorized values recovered; Target: forbidden-value exposure; Extra: any undeclared value; Excess: target recovery beyond the public view.}
	\label{tab:agentleak}
	\setlength{\tabcolsep}{0.5pt}
	\begin{tabular}{lccccc}
		\toprule
		Policy & Recv. & Target & Extra & Excess & Forms \\
		\midrule
		Full context & 1.00 & 1.00 & 1.00 & 1.00 & 1 \\
		Forbidden-list redact & 1.00 & .00 & 1.00 & .00 & 1 \\
		Lexical minimum & .00 & .00 & .00 & .00 & 1 \\
		Fixed abstraction & .00 & .00 & .00 & .00 & 1 \\
		Receipt only & .00 & .00 & .00 & .00 & 1 \\
		\method{} & 1.00 & .00 & .00 & .00 & 1 \\
		\method{}-random & 1.00 & .00 & .00 & .00 & 3 \\
		\bottomrule
	\end{tabular}
\end{table}

We replay the communication layer from AgentLeak commit \texttt{8f0631b9}
\citep{elyagoubi2026agentleak}, holding planning and environment execution fixed while persisting and attacking one sender message. Eligibility requires materialized authorized fields and an auditable forbidden value absent from the public request. Of 150 eligible cases, a seeded domain-balanced subset uses 44 cases each from healthcare, finance, and legal; corporate cases are excluded because required authorized outputs are not materialized. Forbidden-list redaction blocks the named target but exposes other undeclared values, while lexical minimization is incomplete. Both \method{} variants recover all authorized values without observed target or extra exposure.

The randomized renderer uses an independently seeded RNG initialized before dataset parsing; each three-way draw is fixed before the scenario or vault is inspected and recorded in the public view. It uses all forms (41/52/39 cases; 1.573-bit entropy), has zero observed FormGain in the replay audit, and matches deterministic \method{} on utility and exposure. On 480 controlled cases, it obtains receiver success 1.000 and ExcessGain $-.008$ [$-.046$,.029].

We compile planner, sender, receiver, transcript-logger, and attacker nodes in LangGraph 0.4.8. Ten policies over 100 tasks produce 1,000 traces and 5,000 node executions, with actual sender messages passed through graph state, persisted, and attacked. As Table~\ref{tab:langgraph} shows, practical defenses preserve protocol completion but retain transcript leakage; fixed abstraction and receipt only reduce leakage by failing .390 and .680 of requirements. Both \method{} variants pass all evaluated requirements with zero observed ExcessGain. Zero failures in 100 trials give a one-sided exact 95\% upper bound of 2.95\% on the underlying failure probability. This experiment validates executable communication integration rather than long-horizon agent deployment.

\subsection{Schema Coverage and Operational Cost}

Five held-out requirement names are manually mapped to the public-slot grammar. Seen and unseen slot-compatible requirements both achieve receiver success 1.00, near-zero excess gain, and no escalation; raw-evidence requests are rejected with escalation rather than answered by an incomplete abstraction. This evaluates schema compatibility, not learned generalization.

An authenticated-envelope stress test over 900 cases accepts valid requirements and allowlist subsets while rejecting unknown kinds and unauthorized slot expansion. Non-authoritative prompt injections or private-state requests cannot enlarge the authorization object or enter the compiled output. The test assumes structured authenticated inputs and does not evaluate natural-language policy parsing.

The lexical attack preserves the privacy ordering across all three controlled domains and with only 150 attacker-training cases. Removing abstract candidates causes a private-state-aware selector to fall back to fact sharing, whereas the invariant renderer is unaffected because its form is not chosen by leakage scores.

\paragraph{Operational cost.}
Operational \method{} performs policy retrieval, authorized projection, rendering, dependency validation, and an exact rule check without candidate generation, attacker scoring, or LLM calls.

\section{Conclusion}

Privacy-aware structured workflows require control over correlated content and its representation policy.  \method{} formalizes this as a renderer-independent compiler contract; templates are one auditable implementation.  Controlled, AgentLeak, LangGraph, and receiver evaluations show no statistically detectable positive excess gain for \method{} under the evaluated attackers and schemas, while private-state-aware selection leaks through form.  The theorem is narrower: a compliant message adds no information beyond the complete authorized view, which may itself be informative.  Selection invariance is therefore a post-authorization communication-layer principle, not an end-to-end secrecy claim.

\bibliography{aaai2027}

\end{document}